\documentclass[letterpaper, 10 pt, conference]{ieeeconf}  % Comment this line out if you need a4paper
\usepackage{geometry}

\IEEEoverridecommandlockouts                              % This command is only needed if 
\usepackage{bm}
\usepackage{cite}
\usepackage{algorithmic}
\usepackage{graphicx}
\usepackage{textcomp}
\usepackage{xcolor}
\usepackage{algorithm}
\usepackage{wrapfig}
\usepackage{booktabs}
\usepackage{lipsum}
\usepackage{balance}
\usepackage{url}

\let\labelindent\relax
\usepackage{enumitem}
\usepackage{verbatimbox}
\usepackage{tcolorbox}
\usepackage{fancyvrb}
\usepackage{fancyref}
\usepackage{hyperref}
\usepackage{subcaption}
\usepackage[subtle,title=tight]{savetrees}
\usepackage{soul}
\usepackage{amsmath}
\usepackage{amssymb}
\usepackage{amsfonts}
\usepackage{tabularray}
\usepackage{array}
\usepackage[longtable]{multirow}
\usepackage{ragged2e}
\usepackage{longtable}
\usepackage{booktabs}
\usepackage{caption}
\usepackage{multirow}
\usepackage{colortbl}
\usepackage{afterpage}

\newcommand{\hz}[1]{\textcolor{orange}{Zhaohan}}

\usepackage{inconsolata}
\usepackage[T1]{fontenc}

\definecolor{light-gray}{rgb}{0.8, 0.8, 0.8}
\definecolor{comment-green}{rgb}{0.435, 0.576, 0.106}
\definecolor{prompt-blue}{HTML}{2596be}
\definecolor{code-function}{HTML}{379fbe}
\definecolor{code-function}{HTML}{693da8}  % brian (maybe remove)
\definecolor{code-syntax}{HTML}{0060b1}
\definecolor{code-constant}{HTML}{d86001}
\definecolor{prompt-gray}{HTML}{a7a7a7}
\definecolor{highlight}{HTML}{f8f9cb}
\definecolor{highlight}{HTML}{e3eeff}  % brian (maybe remove)
\definecolor{code-perception}{HTML}{2ecc71}
\definecolor{code-control}{HTML}{ff9900}
\definecolor{code-undefined}{HTML}{ff0000}
\renewcommand\fbox{\fcolorbox{light-gray}{white}}

\NewDocumentCommand{\code}{v}{%
\texttt{\small{\textcolor{code-syntax}{#1}}}%
}

\begingroup\lccode`~=`\_ \lowercase{\endgroup\let~}\_
\usepackage{selectp}
\newgeometry{top=72pt, left=54pt, right=54pt, bottom=54pt}
\title{\LARGE \bf
QUART-Online: Latency-Free Multimodal Large Language Model for Quadruped Robot Learning \\
}
\author{
Xinyang Tong$^{1}$†, Pengxiang Ding$^{12}$†, Yiguo Fan$^{1}$†, Donglin Wang$^{1}$*†, Wenjie Zhang$^{1}$, 
Can Cui$^{1}$, Mingyang Sun$^{12}$\\ Han Zhao$^{12}$, Hongyin Zhang$^{12}$, Yonghao Dang$^{3}$, Siteng Huang$^{12}$, Shangke Lyu$^{1}$
\thanks{* means corresponding author and † means co-first authors}
\thanks{$^{1}$MiLAB, Westlake University, Hangzhou, 310030, China}%
\thanks{$^{2}$Zhejiang University, Hangzhou, 310027, China}%
\thanks{$^{3}$Beijing University of Posts and Telecommunications, Beijing, 100876, China}%
\thanks{Email of corresponding author: wangdonglin@westlake.edu.cn}
}
\date{\vspace{-2em}}
\begin{document}

\maketitle

\begin{abstract}

This paper addresses the inherent inference latency challenges associated with deploying multimodal large language models (MLLM) in quadruped vision-language-action (QUAR-VLA) tasks.
Our investigation reveals that conventional parameter reduction techniques ultimately impair the performance of the language foundation model during the action instruction tuning phase, making them unsuitable for this purpose.
We introduce a novel latency-free quadruped MLLM model, dubbed QUART-Online, designed to enhance inference efficiency without degrading the performance of the language foundation model. 
By incorporating Action Chunk Discretization (ACD), we compress the original action representation space, mapping continuous action values onto a smaller set of discrete representative vectors while preserving critical information. 
Subsequently, we fine-tune the MLLM to integrate vision, language, and compressed actions into a unified semantic space.
Experimental results demonstrate that QUART-Online operates in tandem with the existing MLLM system, achieving real-time inference at 50Hz in sync with the underlying controller frequency, significantly boosting the success rate across various tasks by 65\%.
Our project page is \href{https://quart-online.github.io}https://quart-online.github.io.

\end{abstract}
\section{Introduction}

The recent astonishing progress in multimodal large language models (MLLMs) has unveiled their remarkable potential of extracting, aligning, and integrating the representations from complicated language and visual data. 
These advances have driven the development of a versatile quadruped robot, an embodied agent with vision-language understanding and problem-solving skills, adept at interacting with humans and the physical world to perform complex navigation~\cite{tang2023saytap,vinl} and whole-body manipulation~\cite{karnan2022scand} tasks.

An encouraging preliminary work, QUART~\cite{ding2023quar}, has demonstrated the feasibility of adopting MLLMs to control quadruped robots in an end-to-end manner. 
This not only yields performant robotic policies, but also exhibits some emergent abilities obtained from large models, such as understanding novel commands, generalizing to objects never seen before, and reasoning.
Despite these favorable findings, 
the \textbf{high inference latency} is an important bottleneck that inhibits the establishment of generalist quadruped robots with advanced MLLMs.
Typically,
quadruped robotic applications necessitate real-time performance and timely processing for effective action execution and operation in dynamic environments.
However, every time MLLMs are activated to obtain a single robotic action involves a computationally intensive inference process. 
Such inefficiency may prevent real-time integration of perceptual feedback into the decision-making process for actions, thereby leading to the failure of the task.
This weakness makes it challenging to deploy MLLMs on robust quadruped robotic systems.

To alleviate this problem, a straightforward approach is to diminish the quantity of parameters of MLLMs. 
While these methods can reduce inference time, they also compromise the overall performance of the MLLM foundation model, particularly in terms of generalization to unseen scenarios.
In Table~\ref{tab:comparision of reduction methods}, we evaluate QUART with various parameters reduction methods on the QUARD~\cite{ding2023quar} benchmark. 
{While speeding up processing, parameter reduction methods degrade performance, notably in unseen tasks. This suggests a generalization compromise in the foundational capabilities of multimodal models.}
Striking a balance between efficiency and accuracy in model design requires meticulous consideration, prompting the question of whether enhancing performance is achievable without compromising foundational capabilities.
\begin{figure}[t]
    % \captionsetup{type=figure}
    \includegraphics[width=1\linewidth]{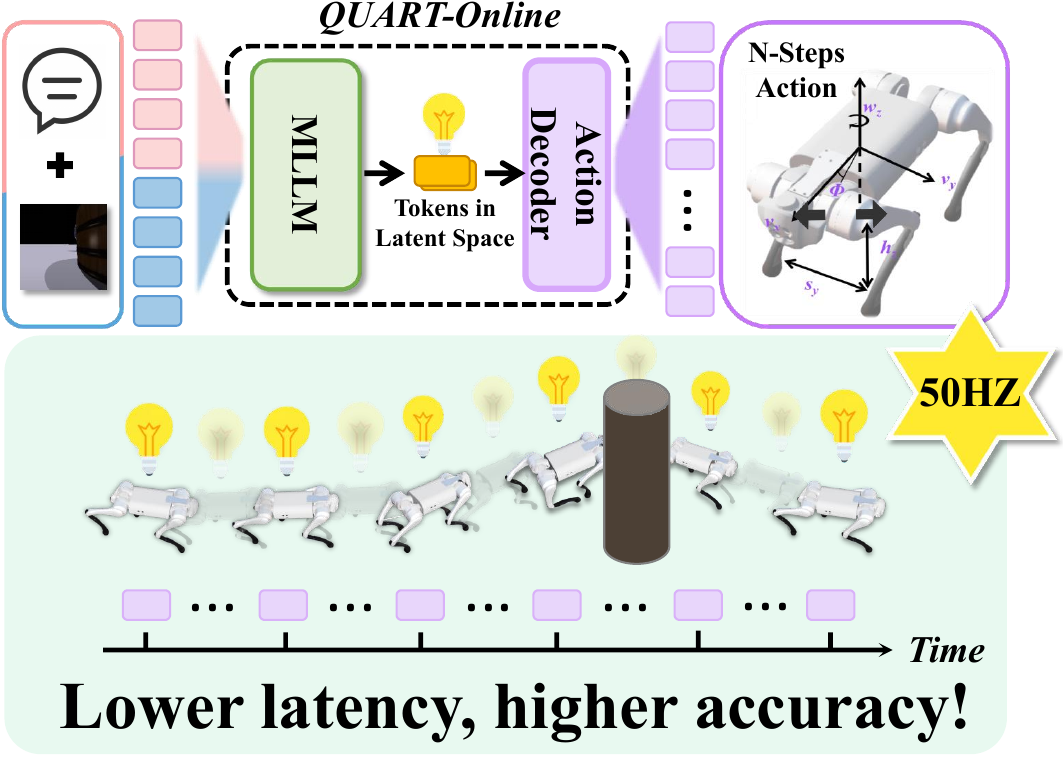}
    \caption{\textbf{Overview of QUART-Online:} With the implementation of action chunk discretization, QUART-Online enhances the existing MLLM system, which was previously operating at a low frequency, enabling more precise actions to be executed in real-time at a frequency of 50Hz.}
    \label{fig:teaser}
\end{figure}

\begin{table}[t]
\caption{{Success rate and inference speed on the unseen multi-task benchmark with various parameter reduction methods on QUART. Here, "P" signifies the implementation of parameter reduction methods on the QUART.}}
\label{tab:comparision of reduction methods}
\renewcommand\arraystretch{1.4} % 行高
% \scriptsize
\centering
\setlength{\aboverulesep}{1.85mm}
\setlength{\belowrulesep}{-0.05mm}
\setlength{\tabcolsep}{1.52mm}{
\begin{tabular}{c|c|c|c}
\hline

{{\textbf{Methods}}} 
&{\textbf{QUART }} 
&{\textbf{QUART \& P }}
&{\textbf{QUART \& P }} 
\\
\hline
{\textbf{Success Rate}}
&0.74  &0.22  &{0.11}\\
\hline
\textbf{Model Parameters }
&8B  &5.3B   &2.7B\\
\hline
\textbf{{Inference Speed}}
&2Hz  &{3Hz}  &{5Hz}\\
\hline
\end{tabular}
}\vspace{-2em}

\end{table}
Thus, this paper introduces QUART-Online, a novel latency-free quadruped multimodal large language model, designed to enhance inference efficiency without degrading the performance of the multimodal large language model. 

\clearpage

\newgeometry{top=54pt, left=54pt, right=54pt, bottom=54pt}
\begin{figure*}[t]
    % \captionsetup{type=figure}
    \includegraphics[width=1\linewidth]{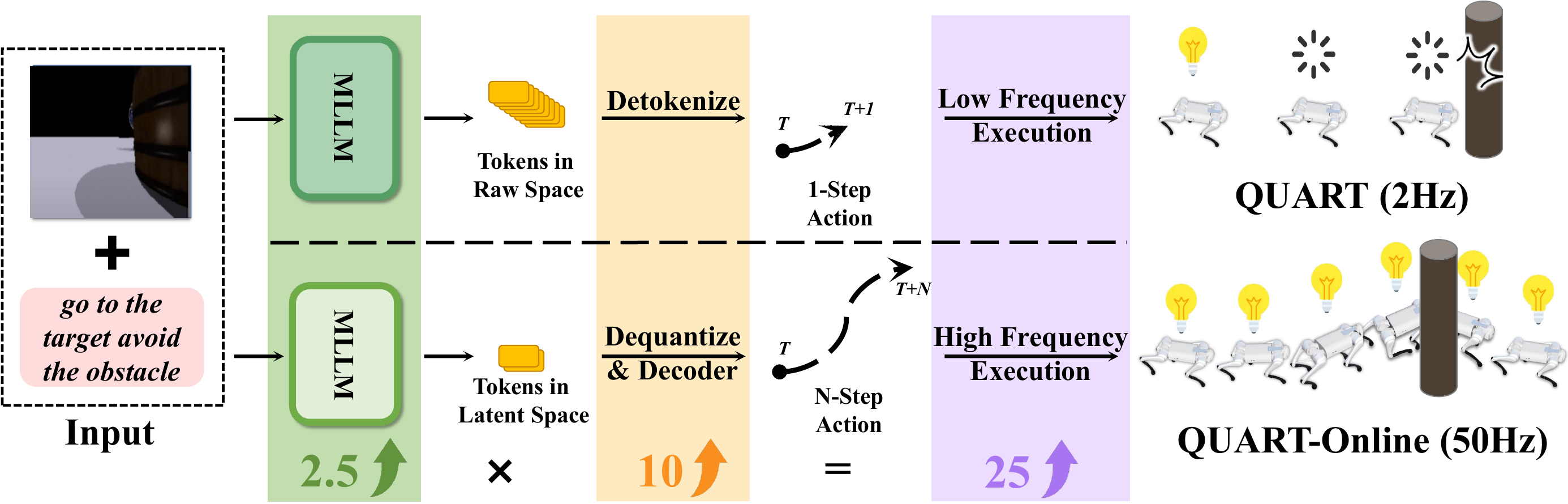}
    \caption{
        \textbf{Comparison of QUART and QUART-Online. }
        QUART-Online enhances the inference process by employing two key strategies: 1) it accelerates MLLM inference by generating a reduced number of tokens in the latent space as opposed to the raw space (2.5x); 2) it introduces an action chunk mechanism during the action decoding phase, facilitating higher-frequency inference via multi-step predictions (10x). By integrating these two approaches, QUART-Online successfully increases the inference rate of the original large quadruped robot model, QUART, from 2Hz to 50Hz, enhancing the model's accuracy in rapidly changing scenarios.
    }
    \label{fig:fig2_overall}\vspace{-2em}
\end{figure*}
Specifically, we first propose a new operation named action chunk discretization (ACD), which maps a sequence of continuous action values onto a smaller set of discrete representative vectors. ACD compresses the original action representation space more effectively while retaining essential information, thereby enhancing the model's output frequency without compromising performance.
Subsequently, we fine-tune the MLLM to integrate vision, language, and compressed actions into a unified semantic space. 
During inference, MLLM first outputs the predicted compressed action tokens, which are then decoded into a continuous trajectory for the robot to execute.

By selecting an appropriate action chunk, QUART-Online aligns with the frequency of the underlying controller, facilitating real-time inference. This precise synchronization ensures swift system responses to inputs, resulting in smoother and more coherent action execution, thereby boosting the system's performance and efficiency in task completion.

The performance of QUART-Online is evaluated on various navigation and whole-body manipulation tasks on QUARD~\cite{ding2023quar} benchmark.
Extensive experiments have demonstrated that QUART-Online when integrated with existing MLLM systems, achieves real-time inference
synchronized with the frequency of the underlying controllers, resulting in a significant 65\% improvement in the average task success rate.

\section{Related Work}

\noindent\textbf{MLLMs for quadruped robot learning.} 
{A range of studies}~\cite{vinl,caluwaerts2023barkour,Yang2021LearningVQ,tang2023saytap,robotics13060086, karnan2022scand, pari2021surprising,10160747,bahl2023affordances,lee2019icra,belkhale2023hydra,haldar2022watch,abk2822,he2024learningvisualquadrupedallocomanipulation,feng2022genlocogeneralizedlocomotioncontrollers,10325606,ouyang2024longhorizonlocomotionmanipulationquadrupedal,jaafar2024lanmp,wu2023learningmultiplegaitslatent,wang2024quadrupedgpt,song2024germ}
have explored the use of natural language to instruct quadruped robots in performing tasks. 
Methods such as SayTap~\cite{tang2023saytap} and QuadrupedGPT~\cite{wang2024quadrupedgpt} utilize LLMs as high-level planners to translate commands into individual primitives that low-level controllers then execute. 
However, these controllers are usually domain-specific small models and lack the semantic understanding and reasoning capabilities that LLMs/MLLMs possess. 
To fully leverage LLM's astonishing capabilities, QUART~\cite{ding2023quar} harnesses the full potential of LLMs by presenting an end-to-end MLLM that generates robotic actions through finetuning on quadruped data, showcasing emergent abilities like a generalization to novel instructions and objects, as well as reasoning.
Despite the potential of projects like QUART in advancing generalist robots, high computational costs often hinder the development of advanced MLLM-equipped quadruped robots, impeding real-time perceptual integration and decision-making. This paper addresses the inference latency challenges in deploying MLLMs for quadruped robot control.

\noindent\textbf{Efficient MLLMs}. 
Significant advancements~\cite{wan2023efficient,han2021dynamic,hinck2024llava,lin2024moe, hinck2024llava, zhao2024cobra, mercat2024linearizing} have been made in enhancing the inference efficiency of MLLMs, such as pruning~\cite{lin2024moe, hinck2024llava}, efficient structures~\cite{mercat2024linearizing}.
However, the aforementioned methods typically require parameter reduction that are made during the pre-training phase, rather than the downstream fine-tuning stage. Forcibly applying these methods during fine-tuning on robot data would inevitably degrade perceptual comprehension capabilities. Thus, this paper aims to find an approach that achieves efficient inference without altering the model structure and maintains perceptual generalization ability.

\noindent\textbf{Robot Action Representation}. 
Robotic motion encoding primarily falls into two categories: {discrete~\cite{brohan2023rt1,brohan2023rt2,belkhale2024rthactionhierarchiesusing,kim2024openvlaopensourcevisionlanguageactionmodel} and continuous~\cite{wu2023unleashinglargescalevideogenerative,li2024visionlanguagefoundationmodelseffective,ke20243ddiffuseractorpolicy,szot2024groundingmultimodallargelanguage}}. Given that large models inherently reason within a discrete space, previous end-to-end approaches~\cite{brohan2023rt1, brohan2023rt2} often employ discrete encoding, mapping all continuous dimensions onto 256 evenly distributed intervals, thereby encapsulating actions as integer values ranging from 0 to 256. This allows the model to align actions with linguistic representations.
This mechanism results in inefficient and suboptimal performance, as the model requires multiple inferences to deduce each action dimension for a single frame, adversely affecting inference speed. Additionally, the model's prediction is limited to the immediate next frame, lacking temporal continuity and leading to unstable and imprecise motion generation.
To address these issues, we propose action chunk discretization (ACD), which allows MLLMs to synchronize real-time inference with the controller's frequency, significantly boosting the success rate in various quadruped robot tasks.

\section{Method}

We begin by outlining the quadruped vision-language-action (QUAR-VLA) task and detailing the first VLA model using MLLM: quadruped robotic transformer (QUART).
Subsequently, we introduce our QUART-Online, a Latency-Free Large Multimodal Language Model for Quadruped Robot Learning from two aspects: \textbf{Action Chunk Discretization} and \textbf{Action Chunk Alignment}.

\subsection{Preliminary}

% 端到端的四足机器人大模型
\noindent\textbf{Quadruped Vision-Language-Action (QUAR-VLA) Task.}
The aim of the QUAR-VLA task is to learn a mapping that interprets RGB images $s \in \mathcal{S}$ alongside task instructions $w \in \mathcal{W}$ to generate future actions through an end-to-end paradigm. 
This task pays particular attention to the connection between text and visual information and robot dynamics instructions and closely relates perception with robot execution.
The mapping is defined as $\mu: \mathcal{S} \times \mathcal{W} \rightarrow \mathcal{A}$, where the action space $\mathcal{A}$ consists of the 11-dimensional high-level commands as well as a terminate signal, i.e.,
\begin{align}
\left[v_x, v_y, \omega_z, \theta_1, \theta_2, \theta_3, f, h_z, \phi, s_y, h_z^f, e \right].
\end{align}

Here, $v_x$, $v_y$, and $\omega_z$ represent the velocities along the x-axis, y-axis, and z-axis respectively. $\theta_1$, $\theta_2$, and $\theta_3$ indicate the gait pattern, $f$ denotes the frequency, $h_z$ represents the height of the robot, $\phi$ denotes the pitch angle, $s_y$ corresponds to the foot width, $h_z^f$ represents the foot height, and $e$ indicates the termination signal of the action.

\noindent\textbf{\textbf{QUA}druped \textbf{R}obotic \textbf{T}ransformer (\textbf{QUART}).} QUART~\cite{ding2023quar} is the first VLA model that leverage an end-to-end MLLM~\cite{BavishiFuyu} to directly generate robotic actions via finetuning on collected quadruped data~\cite{ding2023quar}. 
It takes a single image $s$ and a natural language instruction $w$ as input, which are first converted into concatenated semantic tokens~$c$ through tokenizers~$\tau(c|s,w)$ and fed into a decoder-only transformer module to obtain predicted action tokens~$p(\hat{a_d}|c)$. 
The whole process could be formulated as follows:
\begin{equation}
\begin{aligned}
 &\operatorname{QUART}(\hat{a_c}|s, w) = \operatorname{Detokenize}(\hat{a_d})p(\hat{a_d}|c) \tau(c|s, w),\\
\end{aligned}
\end{equation}
where $w,s$ are the input images and language instruction and $\tau$ represents the tokenizer and $p$ indicates the vision-language model to output action $\hat{a_d}$. Notably, $\hat{a_d}$ still needs a detokenizer to be converted into continuous representation $\hat{a_c}$ for downstream control.

{\subsection{QUART-Online}}

In our pursuit to expedite inference, we initially targeted the most straightforward strategy: model parameter reduction.
Conversely, the performance metrics detailed in Table~\ref{tab:comparision of reduction methods} demonstrate a consistent performance decrement across parameters reduction techniques, with a pronounced decline in efficacy on novel tasks. This trend implies a potential erosion of the intrinsic generalization ability of multimodal models.
This is primarily because architectural optimizations during instruction tuning may inadvertently impair the model's foundation ability, consequently diminishing its performance on targeted tasks.

Therefore, we aim to explore a solution that addresses the inference bottleneck without altering the architecture of the large model or its performance; QUART-Online is our solution.
As shown in Figure.~\ref{fig:fig2_overall}, QUART-Online generates fewer compressed discrete action tokens $\hat{A_q}$ and relies on an action decoder $\mathcal{D}$ to reconstruct $\hat{A_q}$ into continuous action trajectory $\hat{A_c}$: 
\begin{equation}
\begin{aligned}
 & \operatorname{QUART-Online}(\hat{A_c}|s, w) = \mathcal{D}(\hat{A_c}|\hat{A_q})p(\hat{A_q}|c) \tau(c|s, w).\\
\end{aligned}
\end{equation}

Next, we will illustrate the core two components of QUART-Online: 
\textbf{Action Chunk Discretization} and \textbf{Action Chunk Alignment}, where the former maps a sequence of continuous action values onto a smaller set of discrete representative vectors and the latter integrates vision, language, and compressed actions
into a unified semantic space when fine-tuning the MLLM.

{\subsection{Action Chunk Discretization}}

Let's revisit the discretization method initially employed by QUART, which maps continuous dimensions into 256 distinct bins, representing actions as integers ranging from 0 to 255. This approach encounters two primary issues: redundancy in representation and lack of temporal continuity.
Each action frame demands $N$ inference steps, potentially leading to information overlap. Moreover, the action representation fails to maintain temporal consistency and fluidity, causing the generated actions to appear unnatural and disjointed.

To eliminate redundancy, we apply vector quantization to distill raw actions into compressed discrete codes. To ensure temporal coherence, we incorporate it into the learning process of discrete representations, treating multiple consecutive frames as a single chunk. This approach, facilitated by a decoder, allows simultaneous processing of multiple frames, enhancing the fluidity of action generation.

\noindent\textbf{Action Encoder.} To efficiently encode a greater number of temporal action steps into a more compact format, we use $N_L$ layers of 1D convolutions on the temporal axis to compactly encode more action steps, ensuring temporal compression and step coherence. This temporal encoder $\mathcal{E}$ maps action sequences into a high-level latent space, allowing for the extraction of rich semantic and temporal details.
Formally, we treat the sequence of $ N $ consecutive action commands at time $ T $ as a set of input action commands $ {A_c}=\{a^{i}\}_{i=T}^{T+N}\in\mathbb{R}^{N\times 12} $: 
\vspace{-0.3em}
\begin{equation}
\begin{aligned}
& {A_h} = \mathcal{E}(A_c),
\end{aligned}\vspace{-0.4em}
\end{equation}
where encoder $\mathcal{E}$ generates latent action tokens ${A}_{h}\in \mathbb{R}^{C \times D} $, $C$ denotes the compressed temporal dimension, and $D$ represents the dimension of the latent space.

\vspace{-0.3em}
\begin{figure*}[t]
    % \captionsetup{type=figure}
    \includegraphics[width=1\linewidth]{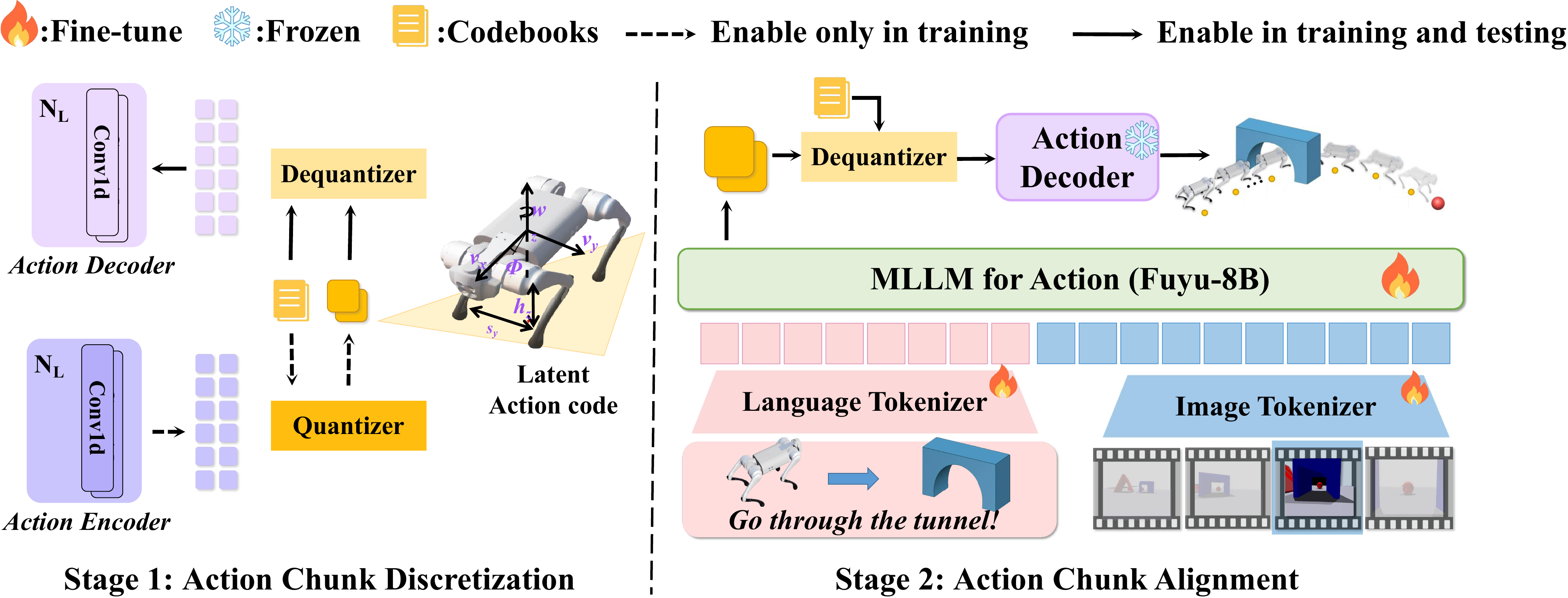}
    \caption{
    {Overall framework of QUART-Online.}
    }
    \label{fig:part1}
    \vspace{-1.5em}
\end{figure*}

\noindent\textbf{Quantization.} 
Notably, our method utilizes a multi-layer quantization, where the input data is sequentially quantized by each quantizer, and the residuals from each quantization step are explicitly managed. 
The learnable codebook $\mathcal{B} \in \mathbb{R}^{N_q \times K\times D}$ consists of $N_q$ layers of codes, where the $i$-th codebook layer are $\mathcal{B}_i = \{b^k\}_{k=1}^K$ including $b^k\in \mathbb{R}^D$ . Here $K$ denotes the number of codes in the codebook, and $D$ represents the dimension of the latent space, corresponding to the number of channels obtained from the encoder.
The quantization iterates $N_q$ times, culminating in the output of the quantized vector ${A}_q \in \mathbb{R}^{C \times N_q}$.
Specifically, during the $i$-th quantization layer, the input vector of the layer is compared with the corresponding codebook layer $\mathcal{B}_i$ to determine the closest match, based on distance calculations. The input vector of each layer is the residual from the previous quantization step, representing the discrepancy between the selected embedding vector and the input vector of that layer. By successively quantifying these residuals, the quantization algorithm refines its approximation of the original high-dimensional data, ensuring encoding accuracy while efficiently managing computational complexity:
% More specific details can be found in the pseudocode representation of \textcolor{red}{Algorithm 1}.
\begin{equation}
{A_q} = \operatorname{Quantize}(A_h, \mathcal{B}).
\end{equation}

\begin{table*}
\centering
% \captionsetup{labelformat=empty}
\caption{We use time per action step (frequency) and the success rate as evaluation metrics, conducting 50 experiments for each task. Here $U_v$ and $U_l$ refer to unseen visual elements and unseen language instructions. The numbers in QUART-Online-1, 5, and 10 represent the lengths of the action chunks.}
\label{tab:overall performance}
\begin{tabular}{c|c!{\vrule width \lightrulewidth}cccc|cccccccc!{\vrule width \lightrulewidth}cc} 
\toprule
\multirow{3}{*}{Method}  & \multirow{3}{*}{Frequency} & \multicolumn{2}{c}{Easy} & \multicolumn{2}{c|}{Medium} & \multicolumn{8}{c|}{Hard} & \multicolumn{2}{c}{\multirow{2}{*}{Average}} \\ 
\cmidrule{3-14}
 &  &   \multicolumn{2}{c}{Distinguish} & \multicolumn{2}{c|}{Go to} & \multicolumn{2}{c}{Go avoid} & \multicolumn{2}{c}{Go through} & \multicolumn{2}{c}{Crawl} & \multicolumn{2}{c|}{Unload} & \multicolumn{2}{c}{} \\ 
\cmidrule{3-16}
 &  &   $U_{v}$ & $U_{l}$ & $U_{v}$ & $U_{l}$ & $U_{v}$ & $U_{l}$ & $U_{v}$ & $U_{l}$ & $U_{v}$ & $U_{l}$ & $U_{v}$ & $U_{l}$ & $U_{v}$ & $U_{l}$ \\ 
\cmidrule{1-2}\cmidrule(r){3-14}\cmidrule(lr){15-16}
\multicolumn{1}{l|}{VLA (CLIP)}  & {43HZ} & 0.40 & 0.56 & 0.66 & 0.27 & 0.53 & 0.22 & 0.31 & 0.06 & 0.00 & 0.2 & 0.01 & 0.00 & 0.32 & 0.22 \\
\multicolumn{1}{l|}{VLA (VC-1)}  & 40HZ & 0.76 & 0.63 & 0.37 & 0.58 & 0.34 & 0.67 & 0.15 & 0.16 & 0.00 & 0.03 & 0.00 & 0.00 & 0.27 & 0.35 \\ 
\multicolumn{1}{l|}{QUART} & 2HZ & \multicolumn{1}{c}{0.80} & 0.89 & 0.59 & 0.76 & {0.2} & 0.41 & 0.23 & 0.40 & 0.32 & 0.46 & 0.08 & 0.18 & 0.37  & 0.52 \\
% \midrule
% \multicolumn{1}{l|}{QUART-Purning-1/3} & & XXHZ & \multicolumn{1}{c}{0.56} & 0.67 & 0.09 & 0.03 & {0.1} & 0.05 & 0.00 & 0.00 & 0.00 & 0.00 & 0.00 & 0.00 & 0.13  & 0.13 \\
% \multicolumn{1}{l|}{QUART-Purning-2/3} & & XXHZ & \multicolumn{1}{c}{0.72} & 0.81 & 0.07 & 0.14 & {0.04} & 0.00 & 0.00 & 0.06 & 0.00 & 0.00 & 0.01 & 0.00 & 0.14 & 0.17  \\
\midrule
\multicolumn{1}{l|}{QUART-Online-1} & 5HZ & \multicolumn{1}{c}{0.48} & 0.75 & 0.25 & 0.76 & {0.05} & 0.33 & 0.33 & 0.68 & 0.56 & 0.5 & 0.00 & \bf{0.42} & 0.28  & 0.57 \\
\multicolumn{1}{l|}{QUART-Online-5}  & 25HZ & \multicolumn{1}{c}{0.82} & 0.99 & 0.61 & 0.90 & {0.39} & 0.54 & 0.26 & 0.42 & 0.48 & 0.59 & 0.06 & 0.28 & 0.47 & 0.62 \\
\multicolumn{1}{l|}{QUART-Online-10}  & 50HZ & \multicolumn{1}{c}{\bf{0.90}} & \bf{0.99} & \bf{0.89} & \bf{1.00} & \bf{0.58} & \bf{0.80} & \bf{0.76} & \bf{0.94} & \bf{0.79} & \bf{0.78} & \bf{0.18} & {0.24} & \bf{0.68} & \bf{0.79} \\
\bottomrule
\end{tabular}
\vspace{-1.35em}
\end{table*}

\noindent\textbf{Dequantization.} The dequantize process is the inverse operation of quantization. Dequantization commences at the apex with layer $\mathcal{B}_{N_q}$ and descends to the foundational layer $\mathcal{B}_1$. Utilizing indices from $\mathcal{A}_q$, it extracts code vectors successively from each layer, aggregating them through a process of accumulation to refine the approximation of the residual incurred at each preceding quantization stage:
\vspace{-0.8em}
\begin{equation}
{\hat{A_h}} = \operatorname{Dequantize}(\hat{A_q},\mathcal{B}).
\end{equation}

\noindent\textbf{Action Decoder.} Decoder $\mathcal{D}$, in contrast to the encoder, takes the dequantized hidden action vector $\hat{A_h}$ and restores it to action in the original action space: 
\vspace{-0.3em}
\begin{equation}
\begin{aligned}
&\hat{A_c} = \mathcal{D}(\hat{A_h}),
\end{aligned}\vspace{-0.5em}
\end{equation}
where $\hat{A_h}\in\mathbb{R}^{N\times 12}$ refers to the action sequence in the original action space that is ultimately reconstructed by the action decoder.

\noindent\textbf{Training Loss.}
The final quantized vector, which is the result of the residual network reconstruction, is denoted by $\hat{A_h}$. The mean squared error between the original $A_h$ and the reconstructed quantized vector $\hat{A_h}$ is used as the commitment loss, denoted as $\mathcal{L}_{\text{com}}$.

% \begin{equation}
% \mathcal{L}_{\text{com}} = \| \mathbf{A_h} - \mathbf{\hat{A_h}} \|^2,
% \end{equation}

% loss
In our approach, we leverage a combination of reconstruction loss and adversarial loss for ACD to achieve high-quality quantization and reconstruction of action tokens:
\vspace{-0.3em}
\begin{equation}
\mathcal{L}_{stage1} = \mathcal{L}_{\text{rec}} +  \mathcal{L}_{\text{com}} = \| {A_c} - {\hat{A_c}} \|^2 + \| {A_h} - {\hat{A_h}} \|^2.
\end{equation}

{\subsection{Action Chunk Alignment}}

As shown in Fig~\ref{fig:part1}, to empower the MLLM to generate action, we need to jointly optimize and align the compressed actions with language and images within the semantic space. 

\noindent\textbf{Training Phase.} 
Upon obtaining a trained vector quantization representation, the target output for the large model transitions from the original discrete values, which span from 0 to 256, to the compressed action tokens. These tokens, in unison with the codebook and the action decoder, facilitate the inference of the final predicted actions.
During this phase, the MLLM undergoes fine-tuning, while the action decoder remains fixed. Given that the final predicted action is a continuous representation, the loss function utilized for this stage follows the cross-entropy loss function employed in the QUART training process:
\vspace{-0.5em}
\begin{equation}
% \text{loss} = -\sum_{i=1}^{N} \sum_{c=1}^{C} y_{i,c} \log(\hat{y}_{i,c})
\mathcal{L}_{stage2} = -\sum A_q \log(\hat{A_q}).
\end{equation}

% \vspace{-0.6em}
\noindent\textbf{Inference Phase.} 
Employing a precisely aligned and concise action chunk representation, we effectively minimize the inference time, thereby enhancing the smoothness and accuracy of the generated actions. 
Additionally, by judiciously selecting an appropriate chunk length, we can achieve latent-free control for large quadruped robot models. This is accomplished when the action chunk length $l_{ac}$ multiplied by the frequency of MLLM inference $f_m$ matches the frequency of low-level control $f_l$. With this well-calibrated chunk length, we can seamlessly implement latency-free control over large quadruped robot models:
\begin{equation}
f_l = l_{ac} \times f_m.
\end{equation}

\section{Experiments}

\begin{table}[t!]
\caption{Reconstruction performance of action chunk discretization. Here ${l_{ac}}$ means the length of the action chunk. }
\label{tab:comparison of reconstruction}
\renewcommand\arraystretch{1.3} % 行高
% \scriptsize
\centering
\setlength{\aboverulesep}{1.85mm}
\setlength{\belowrulesep}{-0.05mm}
\setlength{\tabcolsep}{1.52mm}{
\begin{tabular}{c|c|c|c|c}
\hline

{{\textbf{$l_{ac}$}}}
&{\textbf{MAE(↓)} }
&{\textbf{AKI(↓)}}
&{\textbf{PSNR(↑)} }

&{\textbf{UQI(↑)}}

\\
\hline
\textbf{1 }
 & 0.028 &0.037 &{25.57}  &0.9993\\
\textbf{5}
 & 0.013 &0.0009 &{31.89}  &0.9998 \\
{\textbf{10}} 
 & \bf{0.012} &\bf{0.0008} &\bf{32.11}  &\bf{0.9999}\\
\hline
\end{tabular}
}\vspace{-2.2em}
\end{table}

In this part, we conduct experiments spanning both simulation platforms to address the following inquiries:
1) Can the action chunk discretization (ACD) scheme effectively compress action information?
2) By integrating ACD fine-tuning with MLLM, can QUART-Online maintain model performance while preserving speed?
3) How does the zero-latency inference of QUART-Online enable robots to make better action decisions?
\vspace{-0.5em}
~\subsection{Setup}
\noindent\textbf{Dataset.}
Our experimental analysis employs the QUARD dataset~\cite{ding2023quar}, a comprehensive, large-scale multi-task dataset known as the Quadruped Robot Dataset ({QUARD})~\cite{ding2023quar}. This dataset is rich in content, featuring a variety of tasks including perception, fundamental navigation, and sophisticated capabilities such as whole-body manipulation.
Since the QUARD dataset is collected within NVIDIA's Isaac Gym~\cite{makoviychuk2021isaac}, we chose to use the same simulator as our experimental evaluation environment. 

\noindent\textbf{Evaluation.}
1) Environment: We replicate the QUARD dataset configuration for our evaluation, employing a policy model that outputs high-level commands, which are then translated into robot joint movements by a low-level controller \cite{pmlr-v205-margolis23a}.
2) Tasks: Our analysis extends beyond performance across multiple task scenarios to encompass the model's capability to generalize to novel contexts featuring unseen visual components ($U_v$) and innovative language instructions ($U_l$).
For visual components, we examined items with different textures, colors, and shapes within the same category, as well as those with shape and texture variations not present in current datasets. For language commands, we tested adaptability by using synonymous but differently phrased instructions, such as ``navigate to target'' instead of ``go to the object.''
3) Metrics: We evaluate model accuracy and real-time performance based on success rate and execution frequency. The action compression precision during the ACD phase is quantified using metrics such as mean absolute error (MAE), average kurtosis index (AKI), peak signal-to-noise ratio (PSNR), and universal image quality index (UQI).

\noindent\textbf{Baselines.}
For a thorough comparative analysis, we follow the experimental setup of \cite{ding2023quar, brohan2023rt2}, incorporating the QUART model \cite{ding2023quar} as a paradigm of methods that leverage large language models. Furthermore, we have integrated foundational vision models \cite{radford2021learning, vc2023} with linguistic models to construct distinct vision-language-action (VLA) baselines following the RT-1~\cite{brohan2023rt1} paradigm: VLA (CLIP) and VLA (VC-1).

% \vspace{-0.1em}
\noindent\textbf{{Implementation Details.}} 
In the ACD process, we employ a 1D convolutional Encoder-Decoder with a kernel size of 4 and a stride of 1. The quantization network features a codebook embedding layer with 512 dimensions and 512 quantizers ($N_q=2$). We predict model reconstruction at lookahead steps of N = 5 and N = 10, with 3 convolutional layers at N = 5. The training utilizes AdamW with a learning rate of 3e-4, optimizing the mean squared error loss.
For the multimodal large model, we adopt the lightweight Fuyu-8b as our pre-trained base. Training parameters include a learning rate of 2e-5, AdamW optimizer, 10 epochs, 100K gradient steps, and a batch size of 256. We have addressed previous bugs in the QUART evaluation environment, ensuring all new baselines are reproducible in the updated evaluation setup.

\subsection{Fidelity of action chunk discretization (ACD).}

Given QUART-Online's performance reliance on the quality of ACD, we first assessed the reconstruction accuracy of trajectory discretization. In this study, we allocated 85\% of the action data for training and reserved 15\% for validation.
During the validation phase, we tested three different action chunk lengths (1, 5, and 10) and computed the reconstruction metrics shown in Table~\ref{tab:comparison of reconstruction} to evaluate the fidelity of the trajectory discretization process. The results indicate that as the action chunk length increases, the reconstruction accuracy significantly improves. This enhancement is primarily due to the introduction of additional temporal information, which offers more cues for the reconstruction process.
Furthermore, the results in Table~\ref{tab:overall performance} demonstrate that as the action chunk length increases, QUART-Online's overall performance also shows significant enhancement. This suggests that the increased temporal information improves not only the quality of action reconstruction but also the model’s decision-making accuracy. The following section will elaborate on this in detail.

\begin{figure*}[t]
    \centering
    % \captionsetup{type=figure}
    \includegraphics[width=0.93\linewidth]{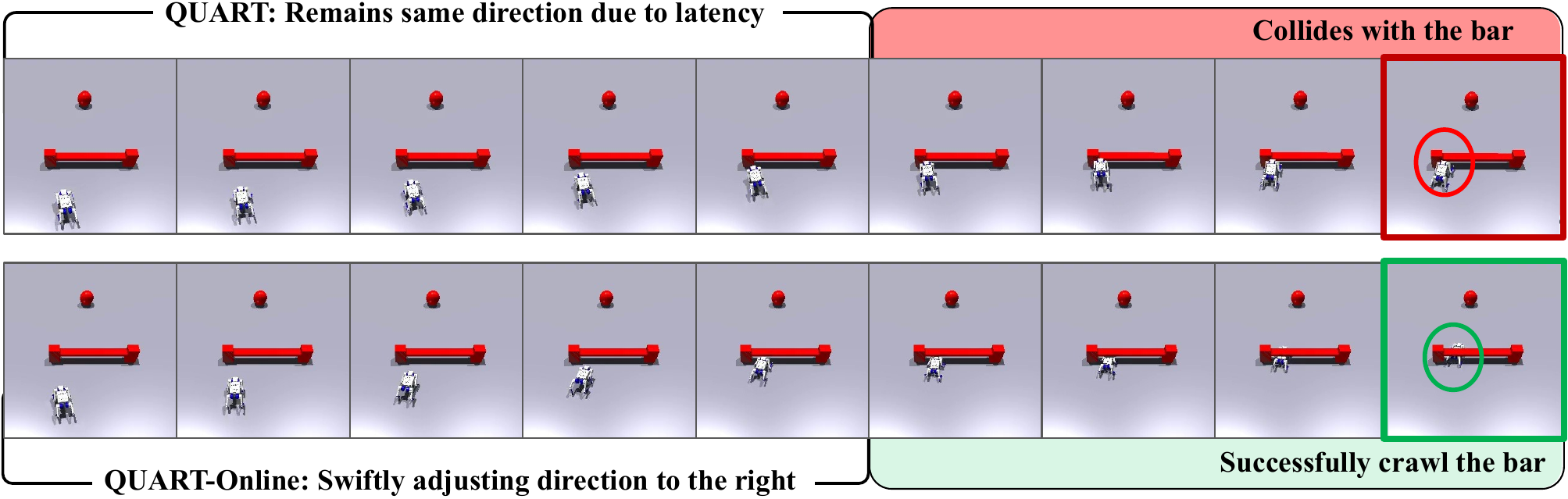}
    \caption{
        {The top half of the comparison highlights the QUART method's latency-induced collision with the red bar (red highlight). The bottom half displays the QUART-Online method's agile response, successfully avoiding the obstacle (green highlight).}
    }
    \vspace{-1.2em}
    \label{fig:visulization}
\end{figure*}
\vspace{-0.6em}
\subsection{Comparison with SOTA methods}
% 我们从两个角度来评价模型的性能：在对于未知物体(obj)和未知语言指令(ins)的泛化性，以及模型的实时响应能力。
We evaluate the model's performance from two dimensions: its
success rate to different scenarios including generalization to previously unseen vision components ($U_v$) and language instructions ($U_l$), as well as its real-time responsiveness. The whole results are shown in Table~\ref{tab:overall performance}.

\noindent\textbf{QUART-Online versus QUART.}
Compared to QUART, QUART-Online has achieved significant improvements in success rates across various tasks. These results demonstrate that QUART-Online significantly enhances multi-task and generalization capabilities compared to QUART, particularly in managing unseen objects and instructions. 
This improvement arises from the unified optimization of actions and perceptual data within a discrete space. Unlike QUART, which compromises semantic distribution by representing actions as discrete numerical tokens aligned with language signals, QUART-Online discretizes actions to extract compressed tokens that convey inherent semantic meaning. This approach allows for joint optimization without disrupting the model's existing distribution, leading to better performance in novel scenarios. Additionally, longer action chunk durations are associated with improved outcomes, suggesting that the encoding of extended time sequences fosters reasoning abilities linked to the current environment, enabling more effective task execution in unfamiliar contexts.
Furthermore, QUART-Online achieves a significant increase in inference frequency, improving from 2Hz to 50Hz. This enhancement enables QUART-Online to operate seamlessly with the existing MLLM system, facilitating real-time inference that aligns closely with the underlying controller's frequency.

\noindent\textbf{QUART-Online versus other baseline models.}
Compared to other baseline models, QUART-Online demonstrates superior performance across all tasks while maintaining a similar inference rate. This reinforces the effectiveness of our approach, which combines the benefits of large models with the real-time capabilities of lightweight models.
Table II illustrates that QUART-Online achieves high success rates, particularly in the "Go avoid" and "Crawl" tasks, outperforming VLA (CLIP) and VLA (VC-1). This combination of high success rates and increased inference frequency enhances the applicability of large models in dynamic robotic tasks.

% \vspace{-1em}
\subsection{Latency-Free Performance Analysis}
% \vspace{-0.5em}

\begin{figure}[t]
    \centering
    % \captionsetup{type=figure}
    \includegraphics[width=0.83\linewidth]{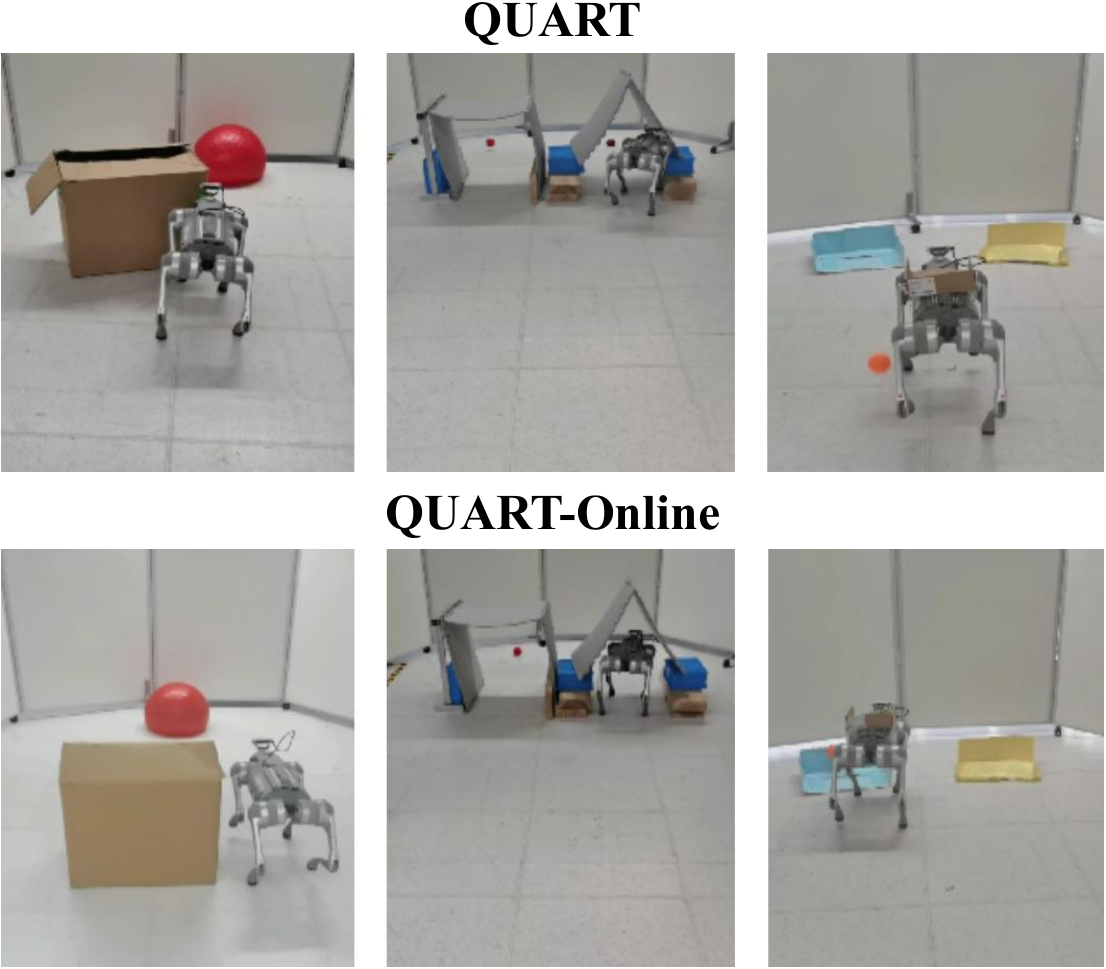}
    \caption{
        {Experiments in the real world.}
    }
    \vspace{-1.7em}
    \label{fig:real_world}
\end{figure}

This part explores strategies for coping with delays in dynamic environments by comparing the performance of QUART and QUART-Online. Figure~\ref{fig:visulization} and Figure~\ref{fig:real_world} illustrate the different responses of the two algorithms when facing the same obstacle (the red horizontal bar). In the case of QUART, the system failed to adjust its direction in time due to delays. As the red ball moved, the robot maintained its original direction, ultimately leading to a collision with the bar (marked by the red box). This result indicates that QUART reacts slowly to dynamic targets and cannot effectively avoid obstacles, highlighting its limitations in real-time applications.
In contrast, QUART-Online demonstrated a more flexible response strategy. This algorithm was able to quickly adjust its direction and timely avoid the bar, successfully passing through (marked by the green box). This shows that QUART-Online has better real-time feedback capabilities and can make more effective decisions in dynamic environments. The experimental results indicate that QUART-Online outperforms QUART under delay conditions, handling dynamic obstacles more effectively. 
% This finding has significant implications for the design of future robotic navigation and control systems, emphasizing the critical role of real-time responsiveness in complex environments.

\section{Discussion and Limitations} \label{sec:discussion}

QUART-Online operates in tandem with the existing MLLM system, achieving real-time inference in sync with the underlying controller frequency, significantly boosting the success rate across various tasks. However, there also exist several limitations:
1) The current actions still do not reach the level of joint angles and still require low-level control to obtain joint angle actions. Therefore, exploring how to directly reach the joint angle layer with a large model in the future is a worthwhile area of research.
2) Our rapid-inferring model has yet to be tested in more complex locomotion scenarios. Ensuring that our efficient inference model can adapt to various complex terrains and further tapping into its potential will be the focus of our next phase of work.

\vspace{-0.5em}
\section{Conclusion}
This paper addresses the challenge of inference latency in MLLMs for controlling quadruped robots. It critiques traditional approaches to parameter reduction, which often lead to a decline in model performance during the downstream tuning phase. In response, the paper introduces QUART-Online, an innovative latency-free method designed to enhance inference efficiency without sacrificing the integrity of the large language model. QUART-Online employs ACD to effectively compress the action space while preserving essential information for precise action mapping.
By fine-tuning with perception data within a unified semantic space, QUART-Online is capable of operating in concert with existing MLLM systems, achieving real-time and accurate inference that is synchronized with the underlying controller's frequency. QUART-Online significantly steps forward in the development of efficient and responsive robotic control solutions for integration with MLLM systems.

\section*{Acknowledgements}
\footnotesize{
This work was supported by the National Science and Technology Innovation 2030 - Major Project (Grant No. 2022ZD0208800), and NSFC General Program (Grant No. 62176215).}
\vspace{-.2em}
% \input{includes/7_ack}


\clearpage
\balance
\bibliographystyle{IEEEtran}
\bibliography{references}

\clearpage

\end{document}